\documentclass[10pt, conference, compsocconf]{IEEEtran}

\usepackage[T1]{fontenc}
\usepackage{amssymb}
\usepackage{graphicx}
\usepackage{todonotes}
\usepackage{latexsym}
\usepackage{algorithm}

\usepackage{algorithmicx}
\usepackage[noend]{algpseudocode}
\usepackage{booktabs}
\usepackage{colortbl,etoolbox,siunitx,xcolor}

\usepackage{subcaption}
\usepackage{amsmath,amssymb}
\usepackage{cite}

\usepackage{listings}
\usepackage{color}
\usepackage[font=footnotesize]{subfig}
\usepackage{url}

\definecolor{dkgreen}{rgb}{0,0.6,0}
\definecolor{gray}{rgb}{0.5,0.5,0.5}
\definecolor{mauve}{rgb}{0.58,0,0.82}

\robustify\bfseries \usepackage{eqparbox}

\newtheorem{problem}{Problem}
\newtheorem{definition}{Definition}

\newcommand{\algName}{\textsc{Gloss}}
\newcommand{\dist}{\operatorname{dist}}
\newcommand{\pdist}{\operatorname{pdist}}
\newcommand{\Expect}{{\rm I\kern-.3em E}}

\DeclareMathOperator\erf{erf}

\begin{document} 

\title{Local Subspace-Based Outlier Detection using Global Neighbourhoods}


\author{\IEEEauthorblockN{Bas van Stein, Matthijs van Leeuwen and Thomas B{\"a}ck}
\IEEEauthorblockA{
LIACS, Leiden University, Leiden, The Netherlands\\
Email: \{b.van.stein,m.van.leeuwen,t.h.w.baeck\}@liacs.leidenuniv.nl}} 


%
%


\bibliographystyle{IEEEtran}
\maketitle

\begin{abstract}
Outlier detection in high-dimensional data is a challenging yet important task, as it has applications in, e.g., fraud detection and quality control. State-of-the-art density-based algorithms perform well because they 1) take the local neighbourhoods of data points into account and 2) consider feature subspaces. In highly complex and high-dimensional data, however, existing methods are likely to overlook important outliers because they do not explicitly take into account that the data is often a mixture distribution of multiple components.

We therefore introduce \algName{}, an algorithm that performs \emph{local subspace outlier detection using global neighbourhoods}. Experiments on synthetic data demonstrate that \algName{} more accurately detects local outliers in mixed data than its competitors. Moreover, experiments on real-world data show that our approach identifies relevant outliers overlooked by existing methods, confirming that one should keep an eye on the global perspective even when doing local outlier detection.
\end{abstract}


\section{Introduction}
\label{Intro}


Outlier detection \cite{hodge2004survey} is an important task that has applications in many domains. In fraud detection, for example, a bank could be interested in detecting fraudulent transactions; in network intrusion detection, it could be of interest to automatically detect suspicious network events; in a manufacturing plant, identifying raw materials or products with strongly deviating properties could be useful as part of quality control. In each of these applications, the data is \emph{high-dimensional} and each \emph{data point} is a potential outlier.

Many techniques for outlier detection have been proposed and studied. Many traditional outlier detection methods \cite{bamnett1994outliers} are parametric and thus make strong assumptions about the data. Moreover, data points are always considered as a whole and relative to all other data points, which strongly limits the accuracy of these methods on high-dimensional data.
Outlier detection in complex, high-dimensional data is an inherently hard problem, as data points tend to have similar distances due to the infamous `curse of dimensionality'. To address both this problem and the limitations of (global) outlier detection, \emph{local} outlier detection methods \cite{Breunig2000,Papadimitriou2003,Kriegel2009} have been proposed over the past few decades. These methods are distance- or density-based, and assign outlier scores based on the distance of a data point to its closest neighbours relative to the local density of its neighbourhood. To further improve on this, local subspace outlier detection methods \cite{keller2012hics,kriegel2009outlier,kriegel2012outlier} have been introduced. They search for local outliers within so-called \emph{subspaces}, i.e., subsets of the complete set of features. This results in each outlier being reported together with a corresponding subspace in which it is far away from its neighbours. Existing local outlier detection approaches, however, are bound to overlook outliers when the data is \emph{a mixture of high-dimensional data points drawn from different data distributions}. That is, as we will show, a local neighbourhood found \emph{within a given subspace} may very well include data points from different components of the mixture, which might result in clear outliers hiding in the crowd of a different component. This is especially relevant when the individual components of the mixture are unknown and hence the dataset has to be analysed as a whole.

We encountered this exact situation in an ongoing collaboration with the BMW Group, where our aim is to identify steel coils strongly deviating in terms of their material properties.
The data is very high-dimensional, as it contains hundreds of measurements per coil, but is also known to be a mixture of samples from different distributions: the steel coils have different grades and come from different suppliers. Unfortunately, part of this information is not available in the data and we therefore had to analyse the complete, mixed data. However, what is a `normal' measurement for one type of coil can be a clear deviation for another type of coil; therefore, existing outlier detection methods did not perform well. Subsection~\ref{subsec:exps:bmw} will show examples of relevant outliers detected by our approach that were not found by existing methods.

\begin{figure}[!tb]
	\centering
	\includegraphics[width=\linewidth]{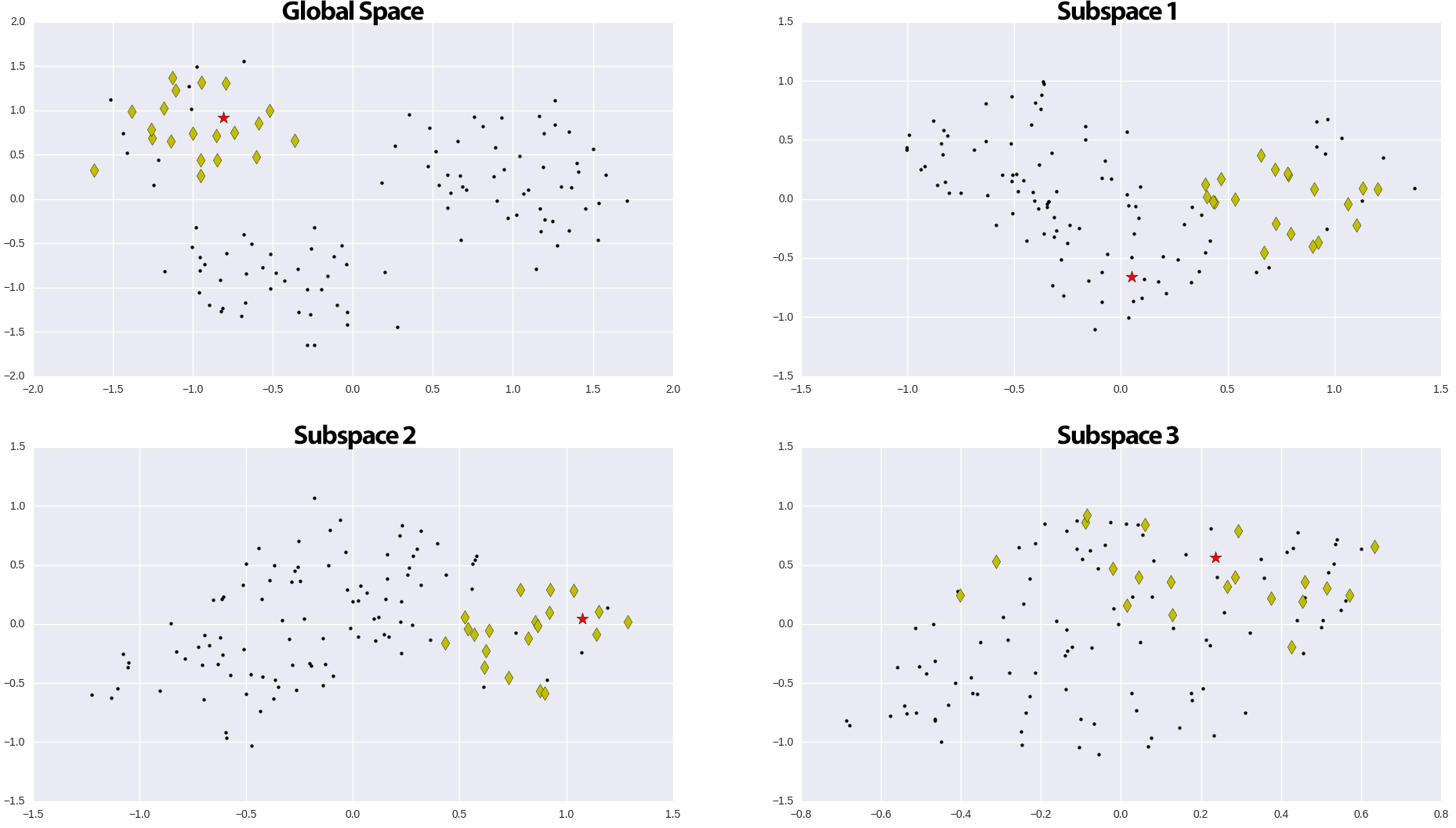}
	\caption{Dataset with six dimensions, consisting of a mixture of samples from three distributions. Shown are the global $6D$ space projected onto $2D$ (top left), and three orthogonal $2D$ subspaces (other). The implanted outlier (red star) can only be detected in Subspace $1$ (top right) if local outlier detection uses its global neighbours (yellow diamonds) instead of its subspace neighbours (dots nearby).\label{fig:toy}}
\end{figure}

Figure~\ref{fig:toy} illustrates the problem that we consider on a synthetic dataset. The data consists of three normally distributed clusters in six dimensions; the generative process (and experiments on generated data) will be described in detail in Subsection~\ref{subsec:exps:synthetic}. When considering all data points, the data point depicted by the red star is not a local outlier in any of the subspaces, neither in the global nor in any of the two-dimensional subspaces (only three shown). However, when only considering \emph{the data point's neighbours in the global space}, here depicted with yellow diamonds, we can observe that the red star is a clear outlier in the 2D subspace shown in the top right plot. \emph{it happens to be close to data points from other components, but is far away from data points from the component it belongs to and is therefore an outlier}. 
As we will show in Subsection~\ref{subsec:exps:synthetic}, existing algorithms are unable to detect such outliers, especially in high-dimensional data, whereas our method can.

\smallskip \noindent 
\textbf{Approach and contributions} 
Our first contribution is the formalisation of the \emph{Local Subpace Outlier in Global Neighbourhood} problem. That is, we propose to combine local subspace outlier detection with neighbourhoods selected in the global data space. The purpose of using global neighbourhoods is to assess the degree of outlierness of a given data point relative to other data points \emph{belonging to the same mixture component}, avoiding the possibility that outliers can hide among members of other components of the mixture distribution.  
Following this, our second contribution is the introduction of the \algName{} algorithm, which combines our ideas on outlier detection using global neighbourhoods with techniques from LoOP \cite{Kriegel2009} and HiCS \cite{keller2012hics}. 

Given a dataset, it computes the probability that a data point is an outlier according to the problem definition. Moreover, it does so for all feature subspaces deemed relevant and hence also provides information about the subspace(s) in which a data point is considered to be an outlier. 

Finally, the third contribution of this paper is an extensive set of experiments on both synthetic and real-world data, in which we evaluate \algName{} and compare its performance to its state-of-the-art competitors. The experiments demonstrate that the use of global neighbourhoods enable the discovery of outliers that would otherwise be left undetected, without sacrificing detection accuracy on `regular' outliers. 
Moreover, global neighbourhoods give \algName{} an edge in terms of computational efficiency. Finally, \algName{} identifies relevant outliers on real-world manufacturing data from the BMW Group that are not marked as such by existing methods. This confirms that outliers can indeed be hidden in mixture distributions in real-world applications and that taking this into account results in better outlier detection.


\section{Related Work}
\label{sec:related}

Although most previous work on outlier detection has been done in statistics, there are also \emph{clustering}-based  \cite{loureiro2004outlier}, \emph{nearest neighbour}-based \cite{hautamaki2004outlier}, \emph{classification}-based \cite{upadhyaya2012classification} and \emph{spectral}-based \cite{choy2001outlier} outlier detection algorithms. Statistical approaches can be categorised as: \emph{distribution}-based \cite{bamnett1994outliers}, where a standard distribution is used to fit the data; \emph{distance}-based~\cite{knorr2000distance}, where the distance to neighbouring points are used to classify outliers versus non-outliers; and \emph{density}-based, where the density of a group of points is estimated to determine an outlier score. While classification, clustering- and distribution-based algorithms aim to find global outliers by comparing each data point to (a representation of) the complete dataset, distance- and density-based algorithms detect local outliers. We next describe the methods most relevant to our paper:

\begin{LaTeXdescription}

\item{Local Outlier Factor} (LOF)~\cite{Breunig2000} was the first algorithm to introduce the concept of \emph{local density} to identify outliers. The authors also claim that they are the first to use a (continuous) `outlier factor' rather than a Boolean outlier class. 

The LOF algorithm uses a user-defined parameter, \emph{MinPts}, that determines the local neighbourhood used for computing the outlier factor for each data point. The outcome of the algorithm strongly depends on this setting. One of the disadvantages of the LOF algorithm is that it is hard to tune the \emph{MinPts} parameter.
Quite some modifications and/or enhancements of LOF, such as the \emph{Incremental Local Outlier Factor} (ILOF) \cite{Pokrajac2007} algorithm, have been proposed. ILOF is a modification of LOF that can handle large data streams and compute local outlier factors on-the-fly. It also updates the profiles of already calculated data points since the profiles may change over time.

\item{Local Correlation Integral} (LOCI)~\cite{Papadimitriou2003} detects outliers and groups of outliers (small clusters) using the \emph{multi-granularity deviation factor} (MDEF). If a point differs more than three standard deviations from the local average MDEF, it is labelled as outlier. 
This method uses two neighbourhood definitions: one neighbourhood 
to use for the average granularity (density) and one neighbourhood 
for the local granularity of a given point. 
The setting of these $\alpha$ and $r$ determines the complexity and accuracy of the algorithm. Typically $\alpha$ is set to $0.5$ and $r$ is set in such way that it always covers at least $20$ neighbours.
\item{Local Outlier Probabilities} (LoOP)~\cite{Kriegel2009} is also similar to LOF but does not provide an outlier factor. Instead, it provides the probability of a point being an outlier using the \emph{probabilistic set distance} of a point to its $k$ nearest neighbours. Given this distance and the distances of its neighbours, a \emph{Probabilistic Local Outlier Factor} (PLOF) is computed and normalised. We will build upon LoOP in this paper.

\item{Subspace Outlier Detection} (SOD)~\cite{kriegel2009outlier} is an algorithm that searches for outliers in meaningful subspaces of the data space or even in arbitrarily-oriented subspaces \cite{kriegel2012outlier}.
Other work in the area of spatial data uses special spatial attributes to define neighbourhood and usually one other attribute to find outliers that deviate in this attribute given its spatial neighbours \cite{chen2010gls,liu2010spatial}.\footnote{More details and a comparison of  these algorithms can be found in \cite{schubert2014local}.}

\item{Outlier Ranking} (OutRank)~\cite{muller2012outlier} determines the degree of outlierness of points using subspace analysis.  For the analysis of subspaces it uses clustering methods and subspace similarity measurements.

\item{High Contrast Subspaces}(HiCS)~\cite{keller2012hics} is a state-of-the-art algorithm that searches for high contrast subspaces in which to perform local outlier detection. It uses LOF as the local outlier detection method for each such subspace, but other algorithms could also be used. 
Runtime is exponential in the number of dimensions, but this can be reduced by limiting the maximum number of subspaces. We will use an adaptation of HiCS for subspace search. 

\end{LaTeXdescription}

Other recent work such as \cite{Campello2015} combines density clustering with local and global outlier detection. We will empirically compare to LOF, HiCS, and two variants of LoOP in Section~\ref{sec:res}, as these are well-studied and representative of the state-of-the-art in local (subspace) outlier detection.


\section{The Problem}

Many outlier detection (and data mining) algorithms assume---either implicitly or explicitly---that the data is an i.i.d. sample from some underlying distribution. That is, assumed is a dataset $D_1$ drawn from some fixed distribution $q_1$, denoted $D_1 \sim q_1$. Given this, global outliers can be found by approximating $q_1$ from the data, estimating $P(d \mid q_1)$ for all $d \in D_1$, and ranking all data points according to the resulting probabilities or scores.

In practice, however, many datasets are mixture distributions of multiple components. Consider for example a dataset $D_2$ consisting of a mixture of two components $C_1$ and $C_2$, drawn from two different distributions, i.e., $D_2 = C_1 \cup C_2$, $C_1 \sim q_1$, and $C_2 \sim q_2$. Globally scoring and ranking outliers now becomes a very challenging task, as identifying the underlying distributions is a hard problem and different components may have different characteristics (such as overall density, attribute-value marginals, etc.).

Local outlier detection algorithms address this problem by considering distances or densities \emph{locally} in the dataset, i.e., within the \emph{neighbourhood} of each individual data point. Although this approach generally works well, it has the disadvantage that it breaks down on high-dimensional datasets, for which all distances become similar; no data points are much further apart than others. 

This problem can be addressed by using a local subspace outlier detection algorithm such as HiCS \cite{keller2012hics}. That is, given a dataset $D$ consisting of data points over a feature space $\mathcal{F}$, these methods search for local outliers within feature subspaces $F \subset \mathcal{F}$. Each reported outlier is associated with a subspace $F$, explaining in which features the data point is different from its neighbours.

However, as argued in the Introduction, this approach suffers from a severe limitation: \emph{existing approaches do not take into account that datasets may be mixtures of multiple components}. That is, when searching for local outliers within a feature space $F$, the density is locally estimated using a neighbourhood determined \emph{using the dataset projected onto the feature subspace only}. Unfortunately, as we will see next, this may have very undesirable side-effects. 

That is, consider again our mixture dataset $D_2$. Suppose that a data point $o \in C_1$, i.e., drawn from $q_1$, is a clear outlier in a (small) subspace $F$, but its values for $F$ are very normal for data points drawn from $q_2$. Then outlier $o$ may go completely undetected by using existing algorithms:
\begin{enumerate}
	\item First, because the data is high-dimensional, global outlier detection methods do not consider $o$ to be far away from other data points in $C_1$ ($o$ is only different in the feature set $F$);
	\item Second, local outlier detection suffers from the same problem when considering all features;
	\item Finally, local subspace outlier detection will not find the outlier either: the neighbourhood of $o$ based on $D_2$ projected onto $F$ \emph{consists of members of component $C_2$}. Although $o$ does not belong to that component, it is in fact very close to those `neighbours' and is therefore not considered an outlier!
\end{enumerate}

Summarising, existing methods cannot detect outliers that 1) are confined to a \emph{feature subspace} but 2) can only be observed within the \emph{global neighbourhood} of the outlier, i.e., when the outlier is compared to data points belonging to the same component. This leads to the following definition.

\begin{problem}[Subspace Outlier in Global Neighbourhood]
	Given a dataset $D$ over features $\mathcal{F}$ and neighbourhood size $k$, we define the probability $p$ that a data point $d \in D$ is a \emph{subspace outlier in global neighbourhood} w.r.t. $F \subseteq \mathcal{F}$ as
\[	
		p_{F,k}(d) = P(\pi_F(d) \mid \pi_F(NN_k(d))),
\]
	where $\pi_F(X)$ denotes $X$ projected onto $F$ and $NN_k(d)$ denotes $d$'s \emph{global} $k$-neighbourhood, i.e., the $k$ data points closest to $d$ in $D$ (over all features $\mathcal{F}$).\label{prob:locglob}
\end{problem}

That is, it is our aim to estimate the probability that a data point is an outlier within a feature subspace, but relative to its neighbours in the complete, global feature space. In the following two sections we will introduce the concepts and theory needed to accomplish this. Note that we will often drop $k$ from $p_{F,k}$ as this is usually a constant.

Before that, however, it is important to observe that we use the global feature space only to determine a reference collection, after which any subspace can be considered for the actual estimation of the outlier probabilities. Although the \emph{absolute} distances between the data points in $\mathcal{F}$ will be small when the data is high-dimensional, a \emph{ranking} of data points based on distances from a given $d$ is likely to result in neighbourhoods that primarily consist of data points belonging to the same component as $d$. That is, we implicitly assume that the components of the mixture are---to a large extent---separable in the global feature space, but this seems very reasonable for the setting that we consider.

\label{sec:problem}

\section{Preliminaries}
\label{sec:prel}

In this section we briefly describe LoOP \cite{Kriegel2009} and HiCS \cite{keller2012hics}, as we will build upon both techniques for our own algorithm, which we will introduce in the next section. The main reason for choosing LoOP is that it closely resembles the well-known LOF procedure but normalises the outlier factors to probabilities, making interpretation much easier. Further, we use an adapted version of the HiCS algorithm to search for relevant subspaces when there is no set of candidate subspaces known in advance.  

\textbf{LoOP} \cite{Kriegel2009} Given neighbourhood size $k$ and data point $d$, LoOP computes the probability that $d$ is an outlier. This probability is derived from a so-called \emph{standard distance} from $d$ to reference points $S$:
\begin{equation}
\sigma(d,S) = \sqrt{ \frac{\sum_{s \in S} \dist(d,s)^2 }{\left|S\right|}  },
\label{eq:sdist}
\end{equation}
where $\dist(x,y)$ is the distance between $x$ and $y$ given by a distance metric (e.g., Euclidean or Manhattan distance).

Then, the \emph{probabilistic set distance} of a point $d$ to reference points $S$ with `significance' $\lambda$ (usually $3$, corresponding to $98\%$ confidence) is defined as
\begin{equation}
\pdist(\lambda,d,S)  = \lambda * \sigma(d,S).
\label{eq:pdist}
\end{equation}
From the following step onward nearest neighbours are used as reference sets. That is, given neighbourhood size $k$ and significance $\lambda$, define the \emph{Probabilistic Local Outlier Factor} (PLOF) of data point $d$ as
\begin{equation}
PLOF_{\lambda,k}(d) = \frac{\pdist(\lambda,d,NN_k(d))}{\Expect_{s \in NN_k(d)} [\pdist(\lambda,s,NN_k(s))] } - 1.
\label{eq:plof}
\end{equation}
Finally, this is used to define Local Outlier Probabilities.

\begin{definition}[Local Outlier Probability (LoOP)]
Given the previous, the probability that a data point $d \in D$ is a local outlier is defined as:
$$LoOP_{\lambda,k}(d) = \max{ \left\{ 0, \erf{\left( \frac{\mathit{PLOF}_{\lambda,k}(d)}{\mathit{nPLOF} \cdot \sqrt{2}} \right) } \right\} } $$
where $\mathit{nPLOF} = \lambda \cdot Stddev(\mathit{PLOF})$, i.e., the standard deviation of PLOF values assuming a mean of $0$, and $\erf$ is the standard \emph{Gauss error function}.
\label{def:loop}
\end{definition}

\textbf{HiCS} \cite{keller2012hics} HiCS is an algorithm that performs an Apriori-like, bottom-up search for subspaces manifesting a \emph{high contrast}, i.e., subspaces in which the features have high conditional dependences. For a given candidate subspace it randomly selects data slices so that a statistical test can be used to assess whether the features in the subspace are conditionally dependent. To make this procedure robust, this is repeated a number of times (Monte Carlo sampling) and the resulting p-values are averaged. Although the method was originally evaluated using both the Kolmogorov-Smirnov test and Welch's t-test, we here choose the former as this does not require any (parametric) assumptions about the data. Parameters are the number of Monte Carlo samples $M$ ($= 50$, default value), test statistic size $\alpha$ ($= 0.1$), and $candidate\_cutoff$ ($= 400$), which limits the number of subspace candidates considered.



\section{The \algName{} Algorithm}
\label{sec:alg}

We introduce \algName{}, for \emph{Global--Local Outliers in SubSpaces}, an algorithm for finding local, density-based subpace outliers in global neighbourhoods, as defined in Problem~\ref{prob:locglob}.
On a high level, \algName{}, shown in Algorithm~\ref{alg:gloss}, employs the following procedure. First, if no subspaces are given a subspace search method is used to find suitable subspaces (Line 1). Then, the global $k$-neighbourhood is computed for each data point in the data (2--3). After that, for each data point an outlier probability is computed for each considered subspace, \emph{relative to its global neighbourhood} (4--9). Finally, these outlier probabilities are returned as result (10). 

As the algorithm computes an outlier probability for each combination of data point and subspace, the probabilities need to be aggregated in order to rank the data points according to outlierness. As we are interested in strong outliers in \emph{any} subspace, we will use the maximum outlier probability found for a data point, i.e., $p(d) = max_{F \in \mathcal{F}}(p_F(d))$. Using the average, for example, would give very low outlier probabilities for data points that [only] strongly deviate in a small subspace.

More in detail, \algName{} builds upon both LoOP and HiCS by integrating both algorithms and adapting them to the global neighbourhood setting that we consider in this paper. The details of outlier detection and subspace search will be described in the next two subsections.

\subsection{Global Local Outlier Probabilities}

First, we introduce the \emph{extended standard distance}, inspired by LoOP, which incorporates 1) a feature subspace $F$ and 2) a \emph{global} neighbourhood relation $G$:
\begin{equation}
\sigma(d_F,G_d) = \sqrt{ \frac{\sum_{s \in G_d} \dist(d_F,s_F)^2 }{\left|G_d\right|}  },
\end{equation}
where $d_F$ and $s_F$ are shortcuts for $\pi_F(d)$ and $\pi_F(s)$ respectively, and $G_d$ is the global neighbourhood defined as $G_d = NN_k(d)$.

Then, using \emph{probabilistic set distance} as defined in the previous section together with the extended standard distance, we define the \emph{Probabilistic Global Local Outlier Factor} $\mathit{PGLOF}$ as:
\begin{equation}
\mathit{PGLOF}_{\lambda,G_d}(d_F) = \frac{\pdist(\lambda,d_F,G_d)}{\Expect_{s \in G_d} [\pdist(\lambda,s,G_s)] } - 1
\end{equation}

Finally, a subspace outlier probability $p_{F,k}(d)$ is computed for each data point and subspace according to Definition 1, but using $\mathit{PGLOF}$ instead of $\mathit{PLOF}$; see Line 9 of Algorithm~\ref{alg:gloss}. That is, with the \emph{global} neighbourhood projected onto the features in the selected \emph{subspace}. 

\begin{algorithm}[!th] 
	\caption{\algName{}}
	\label{alg:gloss}
	\begin{algorithmic}[1]
		\Require Dataset $D$, neighbourhood size $k$, optional: subspaces $\mathcal{F}$
		\State $\mathcal{F} = \operatorname{SubspaceSearch}(D)$ \Comment Only if $\mathcal{F}$ not given
		\ForAll{$d \in D$}:
		\State $G_d = NN_k(d)$
		\EndFor
		\ForAll{$d \in D$}
		\ForAll{$F \in \mathcal{F}$}
		\State $\sigma(d_F,G_d) = \sqrt{ \frac{\sum_{s \in G_d} \dist(d_F,s_F)^2 }{\left|G_d\right|}  }$
		\State $\pdist(\lambda,d_F,G_d)  = \lambda \cdot \sigma(d_F,G_d)$
		\State $\mathit{PGLOF}_{\lambda,G_d}(d_F) = \frac{\pdist(\lambda,d_F,G_d)}{\Expect_{s \in G_d} [\pdist(\lambda,s,G_s)] } - 1$
		
		\State $p_{F,k}(d) = \max{ \left\{ 0, \erf{\left( \frac{\mathit{PGLOF}_{\lambda,G_d}(d_F)}{\mathit{nPGLOF} \cdot \sqrt{2}} \right) } \right\} } $
		\EndFor
		\EndFor\\ 
		\Return $p$
	\end{algorithmic}
\end{algorithm}

\subsection{Subspace Search}

\algName{} can either perform subspace search or use a given set of relevant subspaces. In the latter case, the subspace search (Line 1 in Algorithm \ref{alg:gloss}) is skipped. By parametrising this, we allow background knowledge to be used to reduce the number of subspaces whenever possible, hence avoiding an exponential search for subspaces and thus reducing runtime. In the manufacturing case study that we will present in Subsection~\ref{subsec:exps:bmw}, for example, there is a natural collection of subspaces that can be exploited.

When subspace search is enabled, the search procedure of HiCS is used. However, instead of testing each feature of a candidate subspace against the remaining subspace features, \algName{} tests each candidate subspace feature against the remainder of the \emph{entire feature space}, emphasizing the relation between local and global spaces. As such, the algorithm searches for subspaces that exhibit high contrast relative to the global feature space. Because subspace search is adapted from HiCS, the parameters and their default values are the same as those described in Section~\ref{sec:prel}.


\section{Experiments}
\label{sec:res}

We evaluate \algName{} on 1) synthetic data, 2) benchmark data with implanted outliers, 3) benchmark data with the minority class as outlier class, and 4) a real-world dataset provided by an industrial partner. The source code and experimental setup can also be found on the Github repository\footnote{\algName{} GitHub repository: https://github.com/Basvanstein/Gloss}.

The second and third experiment are available in the preprint version of this paper on arXiv\footnote{For all experiments see the preprint on arXiv: [cs.LG]}.
In the first experiment, in Subsections \ref{subsec:exps:synthetic}, we simulate an (unbalanced) Boolean classification task where the class labels are 1) outlier and 2) not an outlier. This is a very common approach in outlier detection, because objective evaluation is very hard otherwise. Performance is quantified by 1) \emph{Area Under the Curve} (AUC) of the ROC curve and 2) runtime. 

We compare \algName{} to LoOP, LOF, HiCS, and \emph{LoOP local}, a variant of LoOP that detects outliers in each 2D subspace and then assigns the maximum probability over all subspaces to the data point. For all algorithms the neighbourhood size $k$ is set to $20$, which is considered to be sufficiently large; the distance metric is set to Euclidean. For both HiCS and \algName{}, the parameters are set to their defaults and the maximum number of subspaces considered is also set to the default: $100$.

\subsection{Synthetic Data}
\label{subsec:exps:synthetic}

\textbf{Setup} We first devise a generative model to generate data with known outliers that satisfy the assumptions of our problem statement: the data is a mixture of samples from different distributions, and outliers have values sampled from another distribution for some random subspace. More formally, the generative process generates a dataset $D$ with features $\mathcal{F}$ and clusters $C$, where each cluster $c \in C$ is assigned a random center $\mu_c$ and variance ${\sigma^2}_c$.
Each data point $d \in D$ is assigned to one of the clusters uniformly at random, denoted $C(d)$, and then sampled from a normal distribution with specified center and variance:
$$\forall d \in D: d \leftarrow \mathcal{N}({\mu}_{C(d)},{\sigma^2}_{C(d)}).$$

After generating the mixed dataset, outliers $O$ are introduced by changing a random subset of the features for some of the data points. Given a data point $o$, a random $F \subset \mathcal{F}$ and a randomly chosen cluster $r \neq C(o)$, $o$ is marked as outlier and $o$ projected onto $F$ is changed as follows:
$$o_F \leftarrow \mathcal{N}({\mu}_r,{\sigma^2}_r)_F.$$

Experiments are performed on synthetic datasets with $1000$ data points, of which $50$ are marked as outliers. The number of dimensions $d$ is set to $10,20,50,100,200$ or $400$; the number of clusters tested are $2, 3$ and $5$; and $\mu$ is per dimension randomly drawn from $[0,2]$, $[0,3]$, $[0,5]$ or $[0,10]$ (${\sigma^2}_c$ is fixed to $1$). This results in $18$ parameter settings per dimensionality.

\begin{figure*}[!htb]
    \centering

    \begin{subfigure}[t]{0.5\textwidth}%
        \centering
        \includegraphics[width=\textwidth]{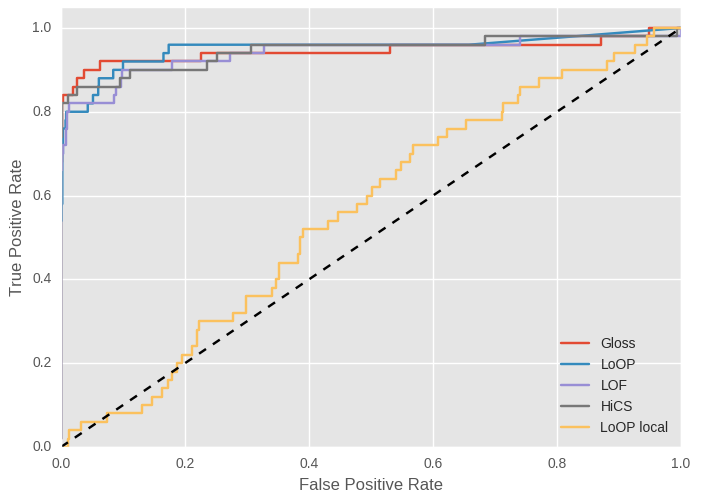}%
        \caption{10D}
    \end{subfigure}%
    ~
     \begin{subfigure}[t]{0.5\textwidth}%
        \centering
        \includegraphics[width=\textwidth]{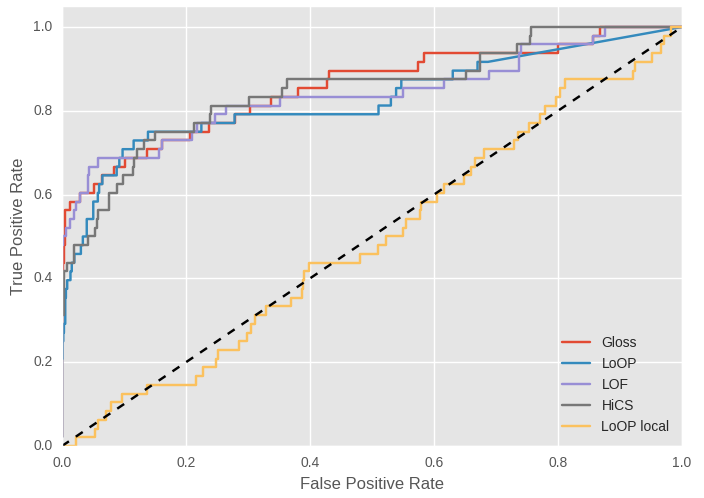}%
        \caption{20D}
    \end{subfigure}%
    
    \begin{subfigure}[t]{0.5\textwidth}%
        \centering
        \includegraphics[width=\textwidth]{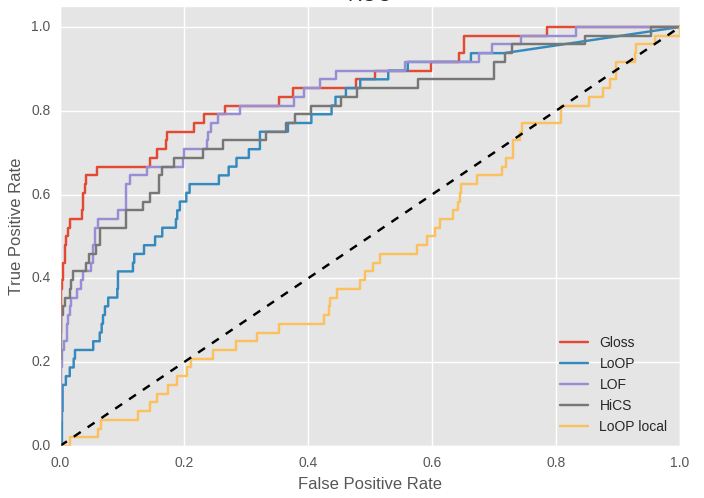}%
        \caption{50D}
    \end{subfigure}%
    ~
    \begin{subfigure}[t]{0.5\textwidth}%
        \centering
        \includegraphics[width=\textwidth]{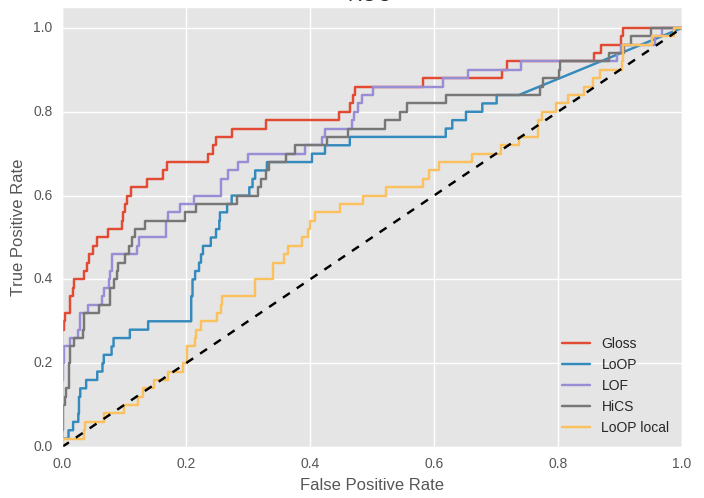}%
        \caption{100D}
    \end{subfigure}%
    
    \begin{subfigure}[t]{0.5\textwidth}%
        \centering
        \includegraphics[width=\textwidth]{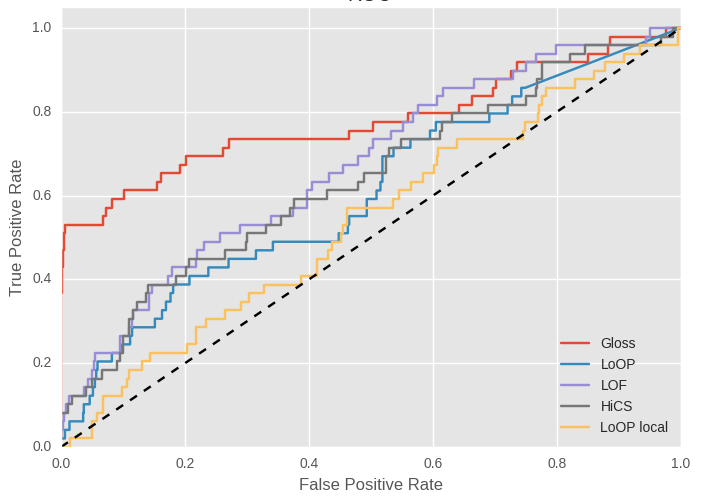}%
        \caption{200D}
    \end{subfigure}%
    ~
    \begin{subfigure}[t]{0.5\textwidth}%
        \centering
        \includegraphics[width=\textwidth]{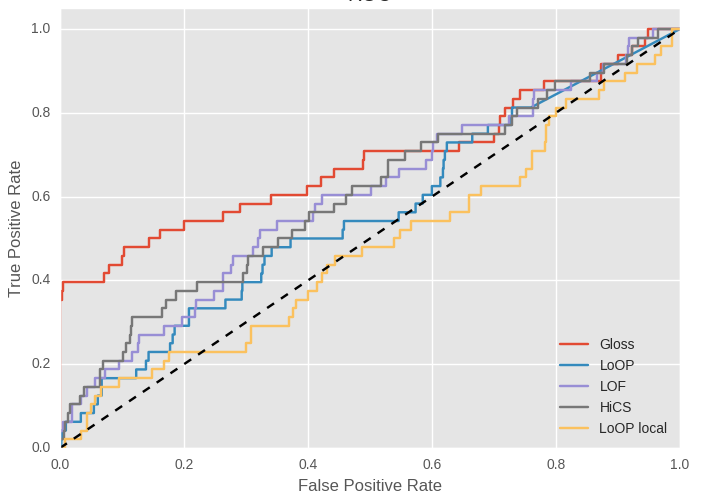}%
        \caption{400D}
    \end{subfigure}%
    \caption{Results on synthetic data. ROC curves for each algorithm and per dimensionality, with $\mu$ randomly drawn from $[0,3]$ and $2$ clusters per dataset.\label{fig:synt}}
    
\end{figure*}

\smallskip \noindent 
\textbf{Results} 
Figure \ref{fig:synt} shows ROC curves for all algorithms per dimensionality, using $3$ clusters and $\mu$ drawn from $[0,3]$. 
Table \ref{table:auc1} presents the obtained AUC scores and runtimes averaged over all $18$ runs per dimensionality. It can be observed from Table \ref{table:auc1} that the purely local subspace analysis done by Local LoOP completely fails to identify the `hidden' outliers, whereas HiCS and the global outlier detection methods fail when the number of dimensions increases. \algName{}, on the other hand, is able to detect most outliers even when the dimensionality increases all the way up to $400$. 
From the ROC curves in Figure \ref{fig:synt} it can be observed that \algName{} tends to find many more outliers at very low false positive rates, while other algorithms only manage to catch up once the false positives rate increases substantially. From the results per individual parameter setting (not included here but available on GitHub), we can see that a higher $\mu$ makes it easier for all algorithms to detect the outliers. This makes sense, since the clusters become more separated and therefore the impact of the local deviation on the global space will be higher. The number of clusters in the data does not seem to be of substantial importance.

\begin{table*}[!htb]
\caption{Results on synthetic data. Average AUC and runtime in seconds for each algorithm,  per dimensionality, averaged over all other parameter settings.}
\label{table:auc1}
\centering
\begin{tabular}{l|c c c c c | r r r r r }

 & \multicolumn{5}{c}{AUC} & \multicolumn{5}{c}{Runtime} \\
\hline
{\ d \ } & {\ \algName{}\ }  & {\ HiCS\ } & {\ LOF\ } & {\ LoOP\ } & {L.LoOP} & {\ \algName{}} & {\ HiCS} &  {\ LOF} & {\ LoOP} & {L.LoOP}  \\
\hline
10 & 0.955 & \textbf{0.964} & 0.959 & 0.956 & 0.547 & 2.28  & 3.21 &  0.05 & 0.44 & 1.93 \\ 
20 & \textbf{0.951} & 0.937 & 0.943 & 0.940 & 0.525 & 4.04 &  3.57   & 0.10 & 0.46 & 3.49 \\
50 & \textbf{0.940} & 0.923 & 0.923 & 0.900 & 0.512 &  12.24 & 9.32 & 0.20 & 0.68 & 8.74  \\ 
100 & \textbf{0.931} & 0.897 & 0.899 & 0.849 & 0.536 &  30.03 & 41.16 & 0.39 & 0.93 & 17.70  \\  
200 & \textbf{0.916} & 0.848 & 0.869 & 0.799 & 0.519 &  79.43 & 91.51 & 0.64 & 1.70 & 38.66  \\ 
400 & \textbf{0.901} & 0.813 & 0.844 & 0.734 & 0.477 &  225.95 & 232.89 &  1.25 & 2.49 & 57.62  \\

\end{tabular}
\end{table*}

\begin{table}[!htb]
\caption{Average execution time for each algorithm per dimensionality setting.}
\label{table:time1}
\centering
\begin{tabular}{l|c c c c c}
{\#D\ } & {\ HiCS\ } & {\ \algName{}\ } & {\ LOF\ } & {\ LoOP\ } & {Local LoOP}  \\
\hline
10 & 3.21 & 2.28  & 0.05 & 0.44 & 1.93 \\           
20 & 3.57 & 4.04  & 0.10 & 0.46 & 3.49 \\          
50 & 9.32 & 12.24 & 0.20 & 0.68 & 8.74  \\        
100 & 41.16 & 30.03 & 0.39 & 0.93 & 17.70  \\  
200 & 336.55 & 79.43 & 0.64 & 1.70 & 38.66  \\  
400 & 2419.99 & 225.95 & 1.25 & 2.49 & 57.62  \\ 
\end{tabular}
\end{table}

\subsection{Benchmark Data with Implanted Outliers}
\label{subsec:exps:benchmark}

\textbf{Setup} We next compare \algName{} to its competitors using a large set of well-known benchmark data from the \emph{UCI machine learning repository} \cite{Bache+Lichman:2013}: Ann Thyroid, Arrhythmia, Glass, Diabetes, Ionosphere, Pen Digits 16, Segments, Ailerons, Pol, Waveform 5000, Mfeat Fourier and Optdigits.

Previous papers usually considered the minority class as `outlier class' for purposes of evaluation, but this clearly would not demonstrate the strengths of our approach: we assume the data to be a mixture of components (i.e., classes), and we search for outliers \emph{within} those classes. We therefore use the UCI datasets as examples of realistic data and implant artificial outliers. That is, we pick a random sample of $10\%$ of the data points and transform each such data point to an outlier by replacing a randomly picked subspace with the values of a data point from a different class (the size of each subspace was chosen uniformly from $[2, max(2, 0.1 * d) ]$). Note that the datasets most likely already contain `natural' outliers, which makes the task at hand even more difficult.


\begin{figure}[!htb]
	\centering
	\includegraphics[width=.5\textwidth]{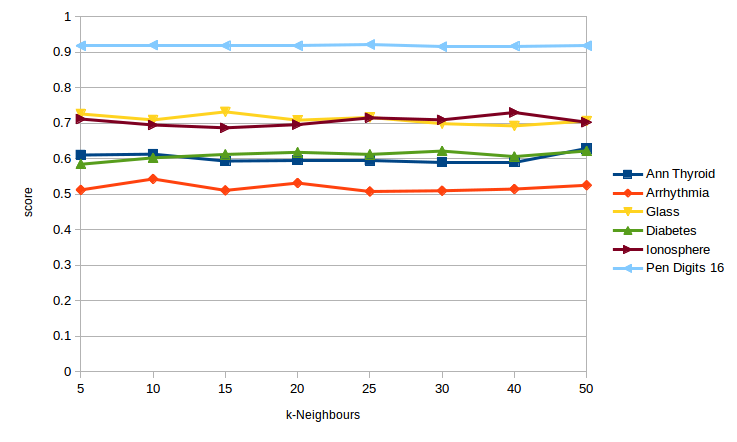}
	\caption{\algName{} accuracy for benchmark datasets with increasing neighbourhood sizes and $5\%$ implanted outliers. \label{fig:increasingk}}
	
\end{figure}


\smallskip \noindent 
\textbf{Results} Figure \ref{fig:increasingk} shows the effect of the neighbourhood size on the performance of \algName{}. It can be seen that the method is very robust with respect to this setting. As $k=20$ is considered to be a ``good'' choice in literature for LoOP and related local outlier detection methods, we chose this as the default for all other experiments and methods.

Table \ref{table:auc2} presents average AUC scores and running times over ten runs per dataset, together with basic dataset properties, for all competing methods. 
\algName{} clearly outperforms its competitors for most datasets when it comes to AUC and is about as fast as HiCS. From this we can conclude that it is beneficial to use \algName{} when the data consists of multiple components and outliers may be hidden as a result of that.


\begin{table*}[!htb]
	\caption{Results on benchmark data with implanted outliers. AUC scores and running times in seconds for each algorithm.}
	\label{table:auc2}
	\centering
	
	\begin{tabular}{l l l|c c c c c | r r r r r }
		
		& \multicolumn{7}{c}{AUC} & \multicolumn{5}{c}{Runtime} \\
		\hline
		{Dataset\ } & {$|D|$} & {$d$} & {\ \algName{}\ }  & {\ HiCS\ } & {\ LOF\ } & {\ LoOP\ } & {L.LoOP} & {\ \algName{}} & {\ HiCS} &  {\ LOF} & {\ LoOP} & {L.LoOP}  \\
		\hline
		\hline
		
		Ann Thyroid  &     3772 & 30 & \textbf{0.608} & 0.565 & 0.595 & 0.591 & 0.545               & 115.851 & 34.864   & 0.316 & 1.259 & 20.11\\
		Arrhythmia  &      452 & 279 & \textbf{0.543} & 0.529 & 0.505 & 0.531 & 0.533               & 327.887 & 1517.813 & 0.134 & 0.471 & 22.447\\
		Glass  &            214 & 9  & \textbf{0.733} & 0.675 & 0.708 & 0.709 & 0.546               & 20.535  & 14.489   & 0.004 & 0.048 & 0.326 \\
		Diabetes &          768 & 8  & \textbf{0.615} & 0.586 & 0.602 & \textbf{0.615} & 0.507      & 25.123  & 10.185   & 0.012 & 0.158 & 0.95\\
		Ionosphere &       351 & 34  & \textbf{0.69}  & 0.608 & 0.636 & 0.61  & 0.573               & 17.65   & 13.171   & 0.015 & 0.098 & 2.074 \\
		Pen Digits 16 &  10692 & 16  & \textbf{0.92}  & 0.86  & 0.915 & 0.91  & 0.497               & 236.956 & 53.43    & 1.555 & 3.07  & 28.759 \\
		Segments         & 2310 & 20 & \textbf{0.815} & 0.8   & 0.765 & 0.797 & 0.745               & 61.859  & 15.292   & 0.159 & 0.595 & 7.619 \\
		Ailerons        & 13750 & 41 & \textbf{0.685} & 0.609 & 0.507 & 0.653 & 0.612               & 396.812 & 91.992   & 0.718 & 8.238 & 109.757\\
		Pol   &           15000 & 49 & 0.576 & 0.538 & \textbf{0.589} & 0.581 & 0.506               & 475.186 & 256.413  & 4.934 & 7.64  & 168.485\\
		Waveform 5000 &    5000 & 41 & 0.568 & 0.568 & \textbf{0.58}  & 0.568 & 0.53                & 172.144 & 76.546   & 2.618 & 2.767 & 34.851\\
		Mfeat Fourier &    2000 & 77 & \textbf{0.627} & 0.543 & 0.594 & 0.567 & 0.512               & 84.378  & 53.707   & 0.797 & 1.232 & 29.244\\
		Optdigits &        5620 & 65 & \textbf{0.688} & 0.584 & 0.675 & 0.659 & 0.518               & 209.598 & 117.749  & 4.951 & 4.601 & 73.045\\

		\hline
		\emph{Average} & & & \textbf{0.672} & 0.622 & 0.639 & 0.649 & 0.552                          & 174.145 & 176.619  & 1.249 & 2.342 & 38.739 \\

	\end{tabular}
\end{table*}

\subsection{Benchmark Data with Minority Class as Outliers}
\label{subsec:exps:benchmark2}

\textbf{Setup} We do not expect using the minority class of a dataset as outlier class to demonstrate the strengths of our approach. Nevertheless, we do not want our improved algorithm to perform worse on the regular local outlier detection task either. Hence, we also compare \algName{} to its competitors using the same benchmark datasets but with outliers defined by the more usual procedure of using the minority class as `outlier class'. Apart from that, we use the same setup and parameters as in Section \ref{subsec:exps:benchmark}.

\smallskip \noindent 
\textbf{Results} Table \ref{table:auc3} presents the average AUC scores obtained over ten runs per dataset.
The results show that \algName{} performs pretty much on par with the state-of-the-art, demonstrating that our proposed method is capable of detecting `regular' outliers as well as the ones that \algName{} identifies but other methods miss (see previous subsections).


\begin{table}[!htb]
	\caption{Results on benchmark data using the minority class as outliers. AUC scores for each algorithm.}
	\label{table:auc3}
	\centering
	
	\begin{tabular}{l |c c c c c  }
		
		{Dataset\ } & {\ \algName{}\ }  & {\ HiCS\ } & {\ LOF\ } & {\ LoOP\ } & {L.LoOP}   \\
		\hline
		Ann Thyroid     & 0.759 & 0.581 & 0.727 & 0.779 & \textbf{0.889} \\
		Arrhythmia      & 0.581 & \textbf{0.646} & 0.48 & 0.617 & 0.582 \\
		Glass           & 0.771 & \textbf{0.818} & 0.815 & 0.744 & 0.621 \\
		Diabetes        & \textbf{0.575} & 0.512 & 0.495 & 0.566 & 0.508 \\
		Ionosphere      & 0.886 & \textbf{0.921} & 0.881 & 0.881 & 0.733 \\
		Pen Digits 16   & 0.473 & 0.522 & 0.461 & 0.465 & \textbf{0.524} \\
		Segments        & 0.522 & 0.493 & 0.512 & 0.52 & \textbf{0.530} \\
		Ailerons        & 0.839 & 0.977 & 0.185 & 0.634 & \textbf{0.987} \\
		Pol             & 0.445 & 0.461 & 0.439 & 0.434 & 0.467 \\
		Waveform 5000   & 0.503 & 0.496 & 0.498 & 0.5 & \textbf{0.512} \\
		Mfeat Fourier   & 0.442 & 0.518 & 0.487 & 0.439 & 0.5 \\
		Optdigits       & 0.519 & \textbf{0.57} & 0.538 & 0.545 & 0.483 \\

		\hline
		\emph{Average}   & 0.61 & \textbf{0.626} & 0.543 & 0.594 & 0.611 \\
	\end{tabular}
\end{table}

\subsection{Case Study: Outlier Detection for Quality Control}
\label{subsec:exps:bmw}

The last series of experiments of this section are performed on a proprietary dataset made available by the \emph{BMW Group} at plant Regensburg. This dataset was one the motivations for this work: the data is high-dimensional and a mixture of different, unknown components. Moreover, it is essential for BMW to be able to identify any outliers in the data, as this directly influences their car manufacturing process.

The data concerns steel coils, which is the raw material used as input at the stamping plant (also called `press shop'). Before entering the stamping process, each coil---of 2--3 km long---is unrolled and cut into shorter pieces. During this process, a large number of measurements is made. We aim to use these measurements to detect steel coils that strongly deviate from a typical coil \emph{in some specific region}. A complicating factor is that the data contains \emph{measurements for different types of steel from different suppliers}, but this important information is not available in the data. Hence, we are dealing with mixed data and we are thus facing exactly the problem formalised as Problem \ref{prob:locglob}, for which we proposed \algName{} as solution.

\smallskip \noindent
\textbf{Setup} The dataset, containing all measurements done from December 2014 to December 2015, consists of $2204$ data points and has $1200$ dimensions, grouped into $100$ $12$-dimensional subspaces using the spatial aspects of the data. Each data point represents a coil having $100$ segments (in length) and $3$ tracks (in width). The most important measurements \cite{purr2015stamping}, and the ones we use, are \emph{Impoc}, quantifying magnetic properties of the steel, and \emph{Oil levels}, quantifying the amount of oil on the coil. Each subspace consists of $3$ Impoc and $9$ Oil level values averaged over a segment of size $2\%$ of the length of the coil; the $100$ subspaces are consecutive, overlapping segments covering the entire coil.

We compare \algName{} to LoOP using all global features and to Local LoOP ran on each of the $100$ individual segments/subspaces. Other algorithms are not included in the evaluation because of the high dimensionality of the data; runtimes would be unreasonably long.

\smallskip \noindent
\textbf{Results} 
As expected, LoOP is unable to detect local outliers: it does not take advantage of the spatial information and cannot deal with the very large number ($1200$) of dimensions. 
The results obtained by \algName{} and our Local LoOP variant are generally similar, but are substantially---and importantly---different for some of the steel coils, as we will show in detail shortly. Moreover, Local LoOP is slower than \algName{}, since the neighbourhood of a coil needs to be computed for each individual subspace, whereas \algName{} only needs to compute a global neighbourhood once.



\begin{figure}[!ht]
    \centering
     \includegraphics[width=.9\linewidth]{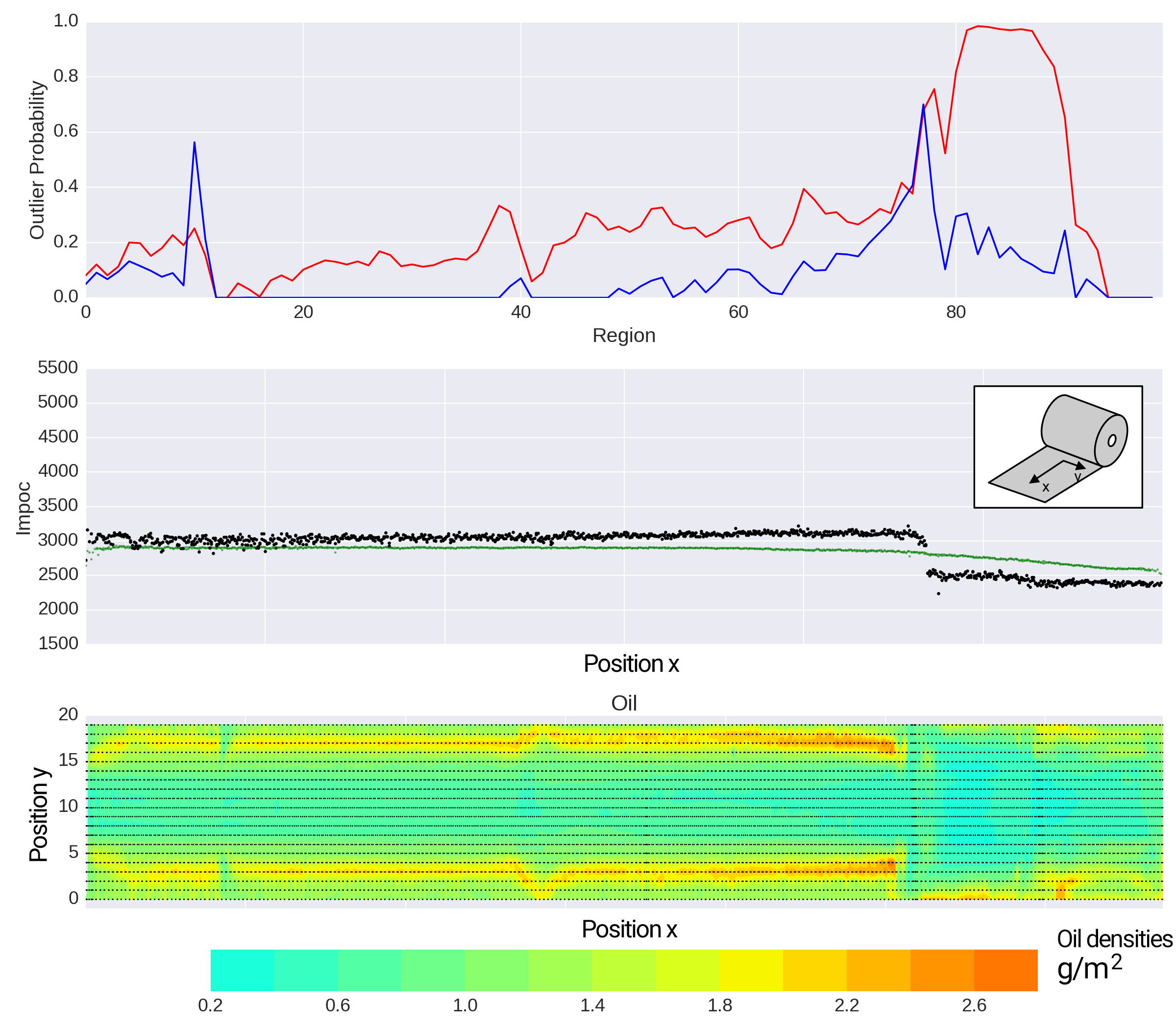}
     \caption{Results for coil $B1$. Top: \algName{} (red line) and LoOP (blue line) outlier probabilities for each of 100 consecutive coil segments. Middle: Impoc measurements over the whole length of the coil, both for this particular coil (black) and averaged over its $20$ global neighbours (green). Bottom: Oil level measurements visualised in 2D, representing the entire surface of the coil.\label{fig:rank287}}
\end{figure}
\begin{figure}[!ht]
    \centering
    \includegraphics[width=\linewidth]{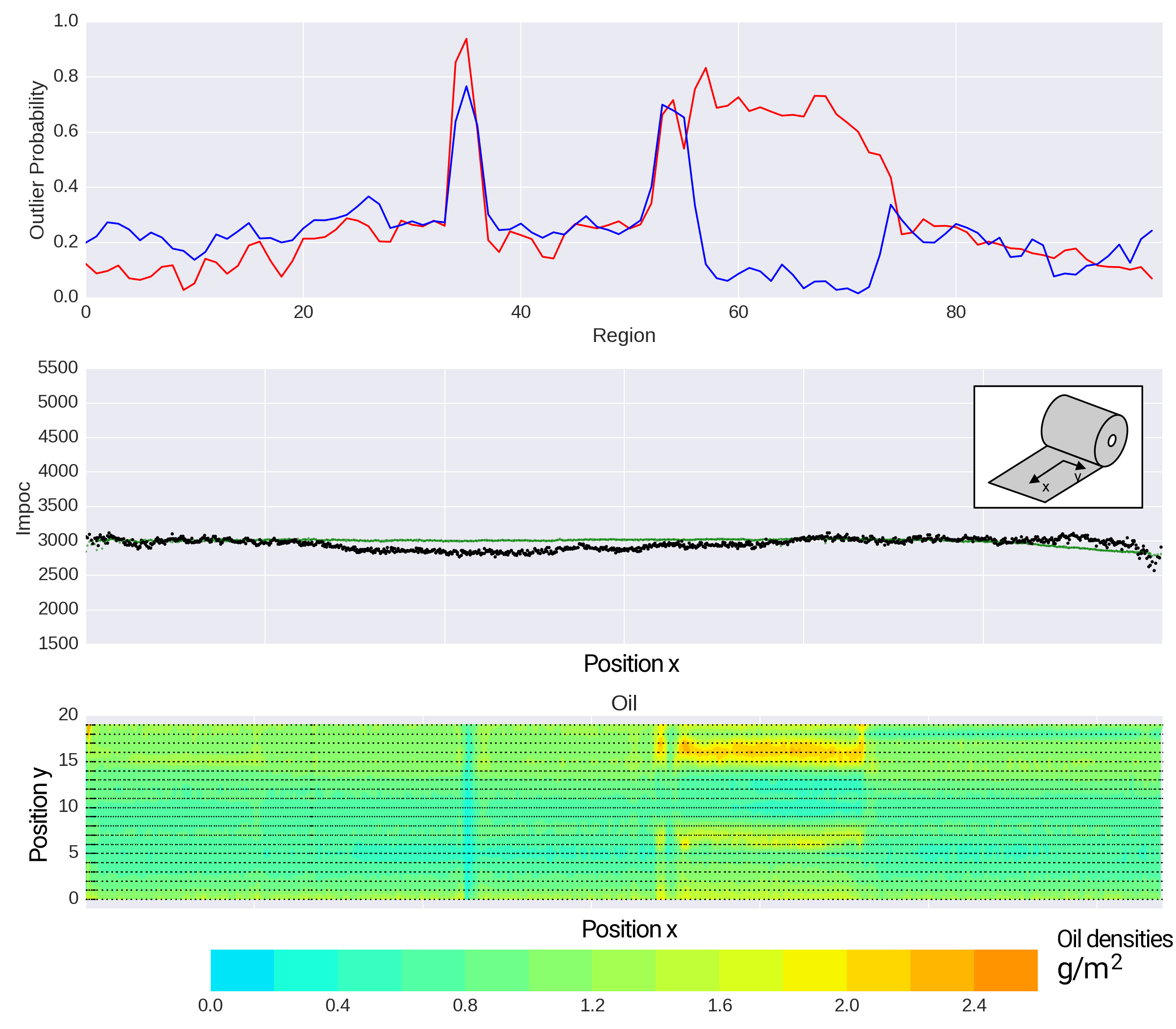}
    \caption{Results for coil $B8$. Presentation details identical to that of Figure \ref{fig:rank287}.\label{fig:rank336}}
\end{figure}

We now zoom in on the $512$ coils recorded in March $2015$, a representative month. By focusing on data from a specific month, we simulate the setting in which the stamping plant operator will inspect the results in the future; \algName{} is currently being implemented in the production environment at BMW. Given that deviations in the steel coils directly influence the manufacturing process, this is expected to improve the stability of the process and the quality of the products.

When comparing the outlier rankings obtained with \algName{} and Local LoOP for this particular month, we observe that many top outliers appear in high positions in both rankings. However, 1) some coils are ranked very differently by the two approaches and 2) \algName{} ranks some coils as outliers that Local LoOP does not. Two such coils are depicted in Figure \ref{fig:rank287} and \ref{fig:rank336}, showing both the outlier probabilities computed by both methods, and the Impoc and Oil level measurements. 
While \algName{} ranks this coil $5$th and $14$th respectively, Local LoOP ranks them $103$th and $79$th. Clearly an operator would inspect this coil, labelled $B1$, if \algName{} were used to rank the coils, but not if Local LoOP would have been used. We asked a domain expert to inspect the measurements and outlier probabilities of this coil and others. He reported back to us that the probabilities computed using \algName{} more accurately reflect the extend to which the coils are outliers.


Next, to further validate the rankings provided by our method, a domain expert of BMW was shown two top-10 outlier coil rankings, one obtained by \algName{} and one by Local LoOP (without  duplicates; a coil was left out from a ranking if it was ranked higher by the other method). Of course, the test was blind, i.e., the domain expert did not know which method generated which ranking. For each coil in either top-10, the domain expert was shown the plots as in Figures \ref{fig:rank287} and \ref{fig:rank336}, but only with the outlier probabilities for the corresponding method. Given the two rankings and plots, the domain expert was asked to rank the $20$ (unique) coils according to the perceived degree of outlierness from the domain perspective. Table \ref{tab:ranking} shows the labels for the coils in the top-10 rankings of Local LoOP and \algName{}, plus the ranking given by the domain expert (using these labels). It is striking that the top four coils selected by the domain expert were all selected by \algName{}, with the top ranked coil being the same coil as the top ranked coil identified by \algName{}.  This confirms that our proposed algorithm is capable of detecting and ranking important outliers that existing algorithms overlook.

\begin{table}[!tb]
\caption{Case study. Outlier rankings obtained by Local LoOP, \algName{}, and BMW domain expert.\label{tab:ranking}}
\centering
\begin{tabular}{l | c c c  }
{Rank\ } &  {L.LoOP} & {\ \algName{}\ }  & {BMW Expert}  \\
\hline
1   &   A1  &   B1  &   B1\\
2   &   A2  &   B2  &   B10\\
3   &   A3  &   B3  &   B6\\
4   &   A4  &   B4  &   B3\\
5   &   A5  &   B5  &   A6\\
6   &   A6  &   B6  &   A1\\
7   &   A7  &   B7  &   B8\\
8   &   A8  &   B8  &   A10\\
9   &   A9  &   B9  &   A4\\
10   &   A10  &   B10  &    B4\\
\end{tabular}
\end{table}

For the application at our industrial partner, deviations in the measurements often indicate problems with the material and these may cause problems during the manufacturing process. Per year, over $100\,000$ coils are processed at this plant, making it infeasible for operators to inspect every single coil. Thus, \algName{} will help to narrow this down by providing outlier rankings and probabilities.


\section{Conclusions}
\label{sec:conc}



Motivated by a real-world problem from the automotive industry, we introduced the generic \emph{Local Subpace Outlier in Global Neighbourhood} problem, and \algName{}, an algorithm that addresses this problem. To enable accurate local subspace outlier detection in high-dimensional data that is a mixture of components, \algName{} uses neighbourhoods selected in the global data space. The experiments show that \algName{} outperforms state-of-the-art algorithms in finding local subspace outliers. Moreover, the experiments show that not only local subspace outliers can be found by \algName{}, but \algName{} performs on par with the state-of-the-art on the regular outlier detection task. The case study on high-dimensional measurement data from steel coils demonstrates that \algName{} is capable at finding relevant local subspace outliers that would otherwise remain undetected, confirming that one should keep an eye on the global perspective even when performing local outlier detection.



\bibliography{main}

\end{document}